\documentclass[runningheads]{llncs}
\usepackage{graphicx}
\usepackage{comment}
\usepackage{amsmath,amssymb} 
\usepackage{color}

\usepackage{epsfig}
\usepackage{graphicx}
\usepackage{amsmath}
\usepackage{amssymb}
\usepackage{enumitem}

\usepackage[pagebackref=true,breaklinks=true,letterpaper=true,colorlinks,bookmarks=false]{hyperref}
\hypersetup{
    colorlinks=true,
    linkcolor=red,
    filecolor=magenta,      
    urlcolor=blue,
}

\newcommand{\improve}[1]{\textcolor{blue}{($\checkmark${#1})}}
\newcommand{\SxTx}[2]{\ensuremath{S_{\times{#1}}T_{\times{#2}}}}

\begin{document}

\pagestyle{headings}
\mainmatter
\def\ECCVSubNumber{132}  

\title{Across Scales \& Across Dimensions: \\
\mbox{\hspace*{-1.25cm} Temporal Super-Resolution using Deep Internal Learning}} 


\titlerunning{Temporal Super-Resolution using Deep Internal Learning}
\authorrunning{Liad Pollak Zuckerman et~al.}
\author{
\small
Liad Pollak Zuckerman\thanks{\hspace*{-0.1cm}joint first authors.}$^1$
\,
Eyal Naor$^{\star 1}$
\,
George Pisha$^{\star 3}$
\,
Shai Bagon$^2$
\,
Michal Irani$^1$
}
\institute{
\small 
$^1$Dept. of Computer Science and Applied Math,
The Weizmann Institute of Science\\
$^2$ Weizmann Artificial Intelligence Center (WAIC)\\
$^3$Technion, Israel Institute of Technology\\
\textbf{\emph{Project~ Website:~}}\url{www.wisdom.weizmann.ac.il/\~vision/DeepTemporalSR}
}

\maketitle

\begin{abstract}
When a very fast dynamic event is recorded with a low-framerate camera, the resulting video suffers from severe motion blur (due to exposure time) and motion aliasing 
(due to low sampling rate in time). 
True Temporal Super-Resolution (TSR) is more than just Temporal-Interpolation (increasing framerate).
It can also recover new high temporal frequencies beyond the temporal Nyquist limit of the input video, thus \emph{resolving both motion-blur and motion-aliasing} -- effects that temporal frame interpolation (as sophisticated as it may be) cannot undo.
In this paper we propose a ``Deep Internal Learning'' approach for  true TSR.
We train a video-specific CNN on examples extracted directly from the low-framerate
input video. 
Our method exploits the strong recurrence of small space-time patches inside a single video sequence, both within and across different spatio-temporal scales of the video. 
We further observe (for the first time) that small space-time patches recur also \emph{across-dimensions} of the video sequence -- i.e., by swapping the spatial and temporal dimensions. 
In particular, the higher spatial resolution of video frames provides strong examples as to how to increase the temporal resolution of that video.
Such internal video-specific examples give rise to strong self-supervision, 
requiring no data but the input video itself.
This results in \textbf{\emph{Zero-Shot Temporal-SR}} of complex videos, which removes both motion blur and motion aliasing, outperforming previous supervised methods trained on 
external video datasets.
\end{abstract}
\section{Introduction}

The problem of upsampling video framerate has recently attracted much  attention  \cite{DAIN,sepconv,nvidia_slomo,Peleg_2019_CVPR,xue2019video,Meyer_2018_CVPR}. These methods perform high-quality Temporal Interpolation on \emph{sharp videos} (no motion blur or motion aliasing). However, temporal-interpolation methods cannot undo motion blur nor  aliasing. 
This is a fundamental difference between Temporal Interpolation and  Temporal Super-Resolution. \\
%

\begin{figure}[!t]
    \centering
    \includegraphics[width=.9\linewidth]{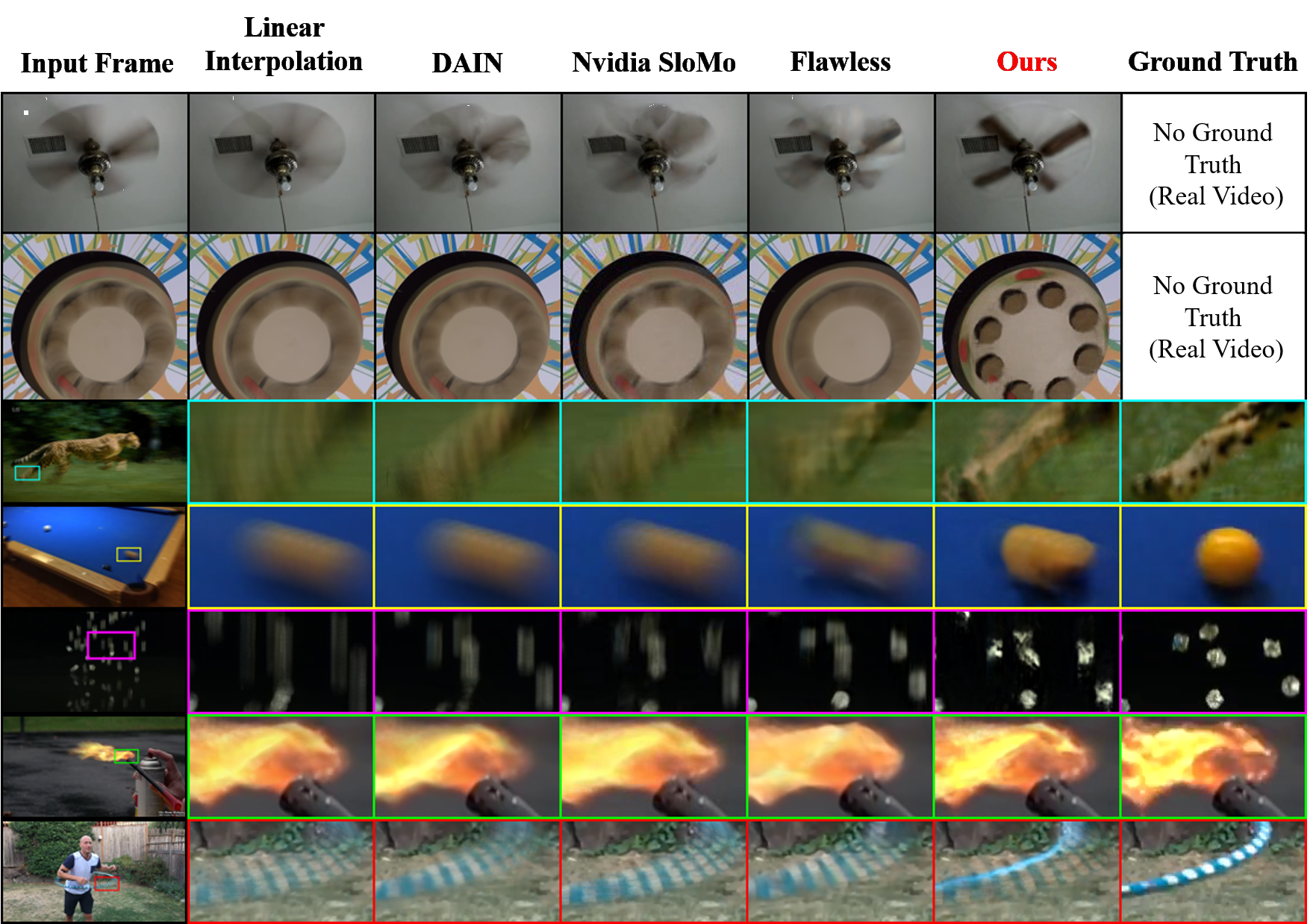}
    \caption{\textbf{Visual Comparison on TS$\times$8.} 
    \emph{\small We compared our method to state-of-the-art methods (DAIN\cite{DAIN},  NVIDIA SloMo\cite{nvidia_slomo}, Flawless\cite{flawless}). Blurs of highly non-rigid objects (fire, water) pose a challenge to all methods. None of the competitors can resolve the  motion blur or aliasing induced by the fast rotating fans. Our {unsupervised} TSR handles these better.
\textbf{View videos in our \href{http://www.wisdom.weizmann.ac.il/\~vision/DeepTemporalSR}{project website} to see the strong aliasing effects}.
}}
    \label{fig:visual-results}
\end{figure}

\begin{figure}[!t]
    \centering
         \includegraphics[width=0.8\linewidth]{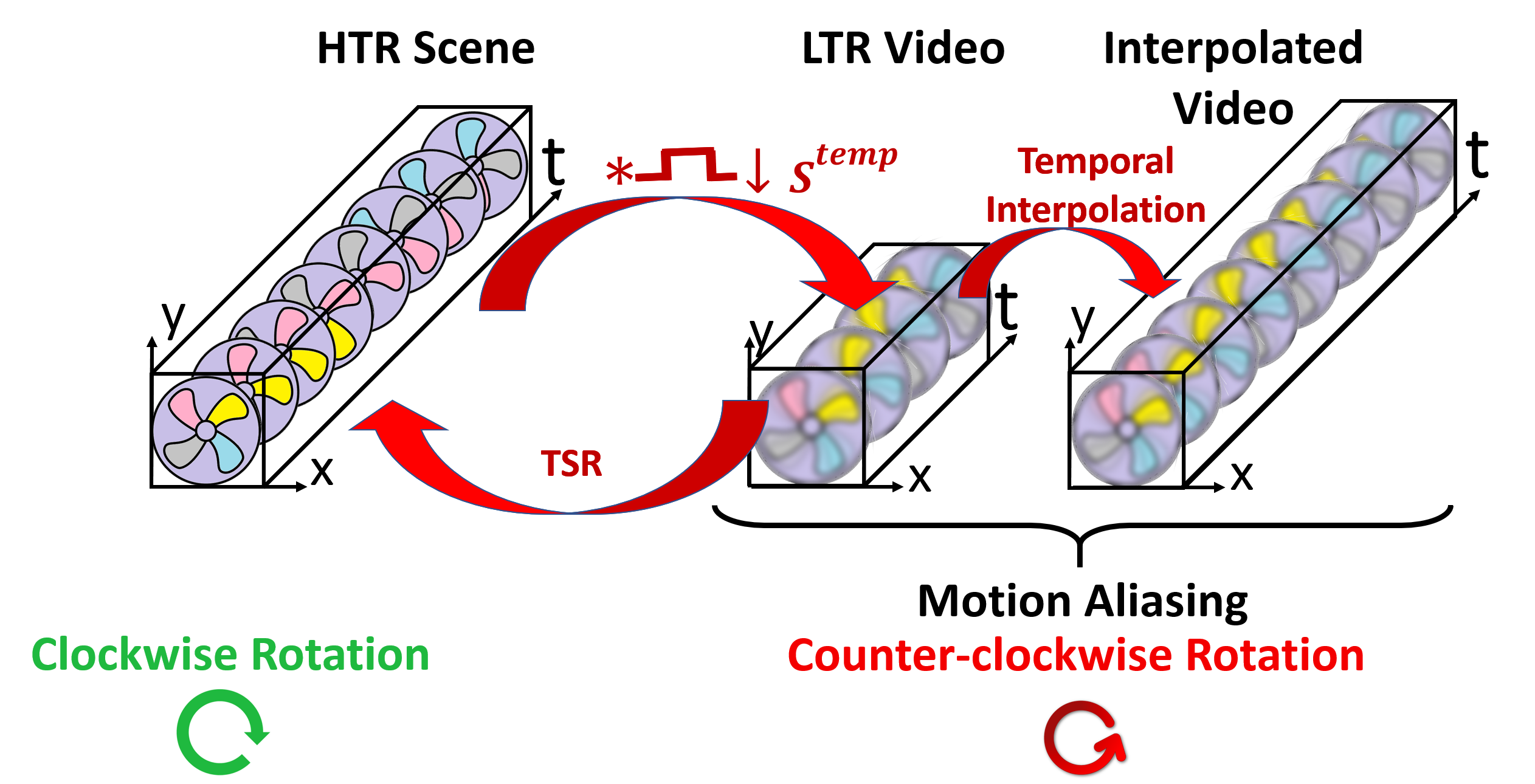}
\vspace*{-0.25cm}
    \caption{
    \textbf{Frame interpolation vs. Temporal-SR.}
    \emph{\small A fan is rotating  \underline{clockwise} fast, while recorded with a 'slow' camera. The resulting LTR video shows a {blurry fan} rotating in the wrong  \underline{counter-clockwise} direction. Temporal frame interpolation/upsampling methods {cannot} undo motion blur nor motion aliasing. They  only add new blurry frames, while preserving the wrong aliased 
    counter-clockwise motion. 
    \textbf{Please 
    see our \href{http://www.wisdom.weizmann.ac.il/\~vision/DeepTemporalSR}{project website} to view these dynamic effects}. In contrast, true TSR not only increases the framerate, but also recovers the lost high temporal frequencies, thus resolving  motion aliasing and blur (restoring the correct fan motion).\vspace*{-.35cm}}
    \label{fig:fan_scene_LTR_interp}}
\end{figure}


\noindent
\textbf{What is Temporal Super-Resolution (TSR)?} 
The \emph{temporal resolution} of a video camera is determined by the frame-rate and exposure-time of the camera. These limit the maximal speed of dynamic events that can be captured correctly in a video. 
\emph{Temporal Super-Resolution} (TSR) aims to increase the framerate in order to unveil rapid dynamic events that occur \emph{faster than the video-frame rate}, and are therefore invisible, or else seen incorrectly in the video sequence~\cite{Shechtman:ECCV02}.

A low-temporal-resolution \textbf{(LTR)} video $L$, and its corresponding high-temporal-resolution \textbf{(HTR)} video $H$, are related by blur and subsampling in time:
\begin{align*}
    L = (H * rect) \downarrow_{{s^{temporal}}}
\end{align*}
where \ $rect$ \ is a rectangular temporal blur kernel induced by the exposure time.

For simplicity, we will assume here that the exposure time is equal to the time between consecutive frames. 
While this is a simplifying 
inaccurate assumption, it is still a useful one, as can be seen in our real video results (Fan video and Rotating-Disk video on our \href{http://www.wisdom.weizmann.ac.il/\~vision/DeepTemporalSR}{project website}). Note that the other extreme -- the $\delta$ exposure model typically assumed by frame interpolation methods~\cite{DAIN,nvidia_slomo}, is also inaccurate. 
The true  exposure time is somewhere in between those two extremes. 


When a very fast dynamic event is recorded with a ``slow'' camera,  the  resulting  video  suffers  from  severe motion  blur   and  motion  aliasing.
Motion blur results from very large motions during exposure time (while the shutter is open), often resulting in distorted or unrecognizable shapes.
Motion aliasing occurs when the recorded dynamic events have temporal frequencies beyond the Nyquist limit of the temporal sampling (framerate). Such an illustrative example is shown in Fig. ~\ref{fig:fan_scene_LTR_interp}. A fan rotating fast \emph{clockwise}, is recorded with a ``slow'' camera. The resulting LTR video shows a \emph{blurry fan} moving in the \emph{wrong direction} --  counter-clockwise.

Frame-interpolation methods~\cite{DAIN,sepconv,nvidia_slomo,Peleg_2019_CVPR,xue2019video,Meyer_2018_CVPR} \mbox{cannot} undo motion blur nor motion aliasing. They  only add new blurry frames, while preserving the wrong aliased counter-clockwise motion  (illustrated in Fig.~\ref{fig:fan_scene_LTR_interp}, and shown for real videos of a fan and a  dotted wheel  in Fig.~\ref{fig:visual-results} \& full videos in the \href{http://www.wisdom.weizmann.ac.il/\~vision/DeepTemporalSR}{project website}). 

Methods for Video Deblurring~(e.g., \cite{kim2015generalized,su2017deep}) were proposed for removing motion blur from  video sequences. These, however, do not increase the framerate, hence cannot resolve  motion aliasing. 

In contrast, true Temporal Super-Resolution (TSR) aims not only to increase the framerate and/or deblur the frames,
%
but also to recover the lost high temporal frequencies beyond the Nyquist limit of the original framerate. 
``Defying'' the Nyquist limit in the \emph{temporal} domain is possible due to the motion blur in the \emph{spatial} domain. Consider two cases: (i)~A fan rotating clockwise fast, which due to temporal aliasing appears to rotate slowly counter-clockwise; and (ii)~A fan rotating slowly counter-clockwise. When the exposure time is long (not $\delta$), the fan in (i) has severe motion blur, while the fan in~(ii) exhibits the same motion with no blur. Hence, while temporally (i) and (ii) are indistinguishable, spatially they are.
Therefore, TSR can resolve both motion-aliasing and motion-blur,
producing sharper frames, as well as the true motion of the fan/wheel (clockwise rotation, at correct speed). 
See  results in Fig.~\ref{fig:visual-results}, and full videos in the \href{http://www.wisdom.weizmann.ac.il/\~vision/DeepTemporalSR}{project website} 
(motion aliasing is impossible to display in a still figure).

A new recent method, referred to in this paper as `Flawless'~\cite{flawless},
presented a Deep-Learning approach for performing \emph{true TSR}. It trained on an external dataset containing video examples with motion blur and motion aliasing. Their method works very well on videos with similar characteristics to
their training data -- i.e.,  strong camera-induced motion blur on mostly rigid scenes/objects. However, its performance deteriorates on natural videos of more complex dynamic scenes -- highly non-rigid motions and severe motion-aliasing (see Fig.~\ref{fig:visual-results}). 

Failure to handle non-typical videos is not surprising –- generating an inclusive video dataset containing  all possible combinations of spatial appearances, scene dynamics, different motion speeds, different framerates, different blurs and different motion aliasing, is combinatorially infeasible. 

\begin{figure}[!t]
    \centering
    \includegraphics[width=0.85\linewidth]{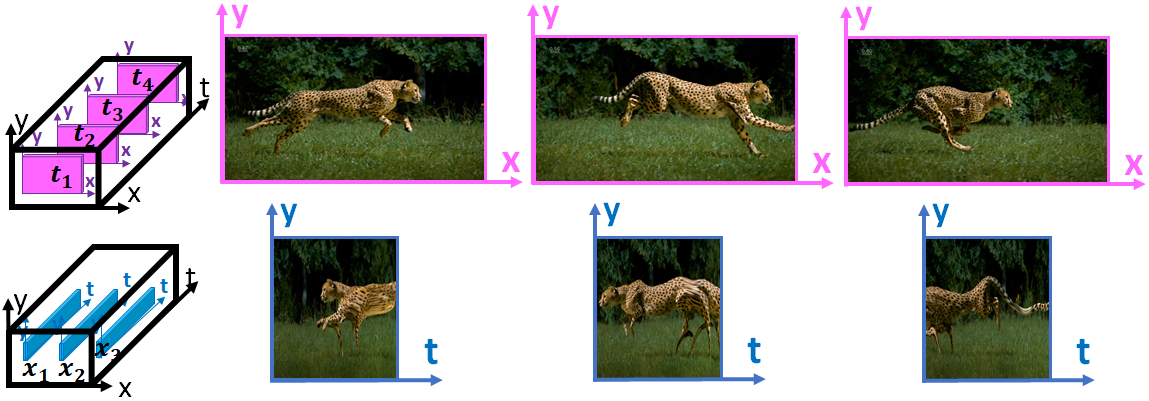}
    \caption{\textbf{Slices of the space-time video volume.} \emph{\small (Top:)~xy slices = frames,  \  (Bottom:)~ty~slices. The xy slices (video frames) provide high-resolution patch examples for the patches in the low-resolution ty slices. (See text for more details)
    }}
    \label{fig:All_slices}
\end{figure}
\begin{figure}[!t]
\vspace*{-0.1cm}
    \centering
    \includegraphics[width=0.8\linewidth]{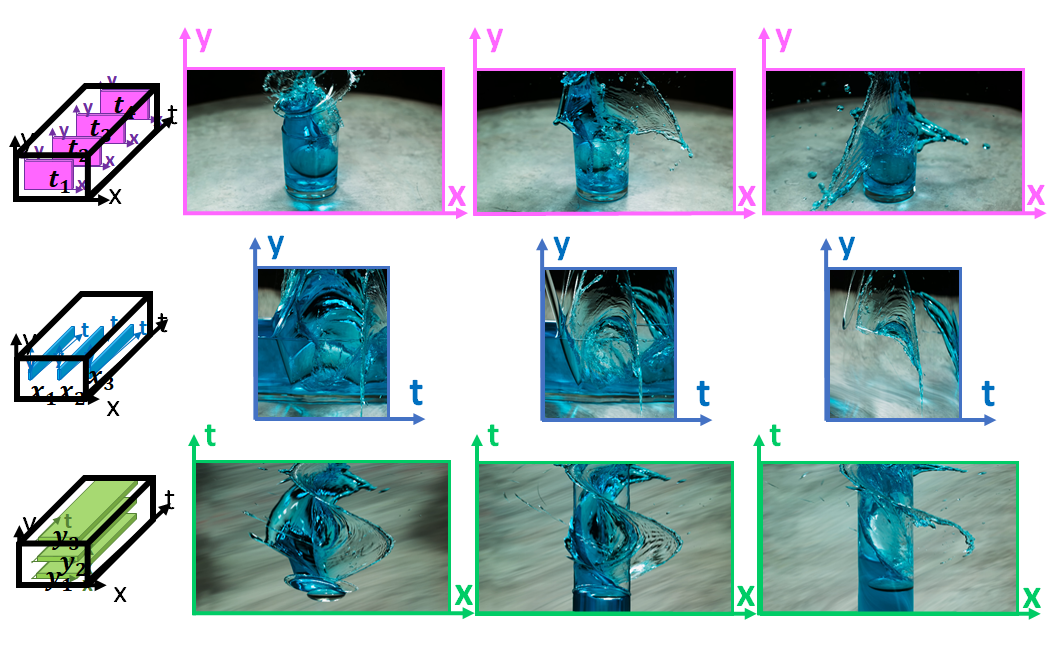}
      \vspace*{-0.25cm}
    \caption{\textbf{Slices of the space-time video volume of a complex  motion.} \emph{\small (Top to bottom:) xy slices, ty slices, xt slices. Note that  patch similarities across-dimensions hold even for highly complex non-linear motions. 
   See \href{http://www.wisdom.weizmann.ac.il/\~vision/DeepTemporalSR}{project website} for videos of slices across different dimensions.
   \vspace*{-0.35cm}
}}
    \label{fig:glass_slices}
\end{figure}

In this work we propose to overcome these dataset-dependant limitations by replacing the External training with Internal training. Small space-time video patches have been shown to recur across different spatio-temporal scales of a single natural video sequence~\cite{Shahar2011STSR}. This strong internal video-prior was used by~\cite{Shahar2011STSR} for performing TSR from a single video (using Nearest-Neighbor patch search within the video). Here we exploit this property for training a Deep Fully Convolutional Neural Network (CNN) on examples extracted \emph{directly from the LTR input video}. We build upon the paradigm of ``Deep Internal Learning'', first coined by~\cite{ZSSR_a}. They train a CNN solely on the input image, by exploiting the recurrence of small image-patches across scales in a single natural image~\cite{glasner}. This paradigm was successfully used for a variety of image-based applications~\cite{ZSSR_a,ingan,singan}. 
Here we extend this paradigm, for the first time, to video data. 


We further observe (for the first time) that \emph{small space-time patches (ST-patches)  recur also \textbf{across-dimensions} of the video sequence}, i.e., when swapping between the spatial and temporal dimensions (see Fig.~\ref{fig:All_slices}).
In particular, {the higher \emph{spatial resolution} of video frames provides strong examples as to how to increase the \emph{temporal resolution}} of that video (see Fig.~\ref{fig:Training_set}.b). We exploit this recurrence of ST-patches across-dimensions (in addition to their traditional recurrence  across video scales), to generate \emph{video-specific training examples}, extracted directly from the input video. 
These are used to train a video-specific CNN, 
%
resulting in \emph{\textbf{Zero-Shot Temporal-SR}} of complex videos, which resolves both motion blur and motion aliasing. It can handle  videos with complex dynamics and highly non-rigid scenes (flickering fire, splashing water, etc.), that supervised methods trained on external video datasets cannot handle well. 

\vspace*{0.1cm}
\noindent
\textbf{Our contributions are several-fold:}
\begin{itemize}[noitemsep,nolistsep,leftmargin=*,label=$\bullet$]
\item Extending ``Deep Internal Learning'' to video data.
\item Observing the recurrence of data \emph{across video dimensions} (by swapping space and time), and its implications to TSR.
\item Zero-Shot TSR (no training examples are needed other than the input video).
\item We show that internal training resolves motion blur and motion aliasing of complex dynamic scenes,  better than externally-trained  supervised methods.
\end{itemize}
%

\section{Patch Recurrence across Dimensions}
\label{sec:across_dim}
%
%
It was shown~\cite{Shahar2011STSR} that small Space-Time (ST) patches tend to repeat abundantly inside a video sequence, both within the input scale, as well as across coarser spatio-temporal video scales. Here we present a new observation \emph{\textbf{ST-patches recur also across video dimensions}}, i.e., when the spatial and temporal dimensions are swapped. Fig.~\ref{fig:All_slices} displays the space-time video volume (\mbox{x-y-t}) of a running cheetah. 
The video frames are the \emph{spatial} x-y slices of this volume (marked in magenta). Each frame
corresponds to the plane (slice) of the video volume at time $t$$=$$t_i$. Swapping the spatial and the temporal dimensions, we can observe ``frames'' that capture the information in y-t slices ($x$$=$$x_i$ plane) or x-t slices ($y$$=$$y_i$ plane). Examples of such slices appear in Figs.~\ref{fig:All_slices} and~\ref{fig:glass_slices} (green and blue slices).
These slices can also be viewed dynamically, by flipping the video volume (turning the \mbox{x-axis} (or \mbox{y-axis}) to be the new t-axis), and then playing as a video. Such examples are found in the \href{http://www.wisdom.weizmann.ac.il/\~vision/DeepTemporalSR}{project website}.

When an object moves fast, patches in x-t and y-t slices appear to be \emph{low-resolution versions} of the higher-resolution x-y slices (traditional frames). Increasing the resolution of these x-t and y-t slices in t direction is the same as increasing the temporal resolution of the video. The \emph{spatial} x-y video frames thus provide examples as to how to increase the \emph{temporal} resolution of the x-t and y-t slices within the same video.  Interestingly, when the object moves very slowly, patches in x-t and y-t slices appear as stretched versions of the patches in x-y frames, indicating that these temporal slices may provide examples as to how to increase the \emph{spatial} resolution of the video frames. This however, is beyond the scope of the current paper.

\begin{figure}[!t]
\vspace*{-0.25cm}
    \centering
    \includegraphics[width=0.7\linewidth]{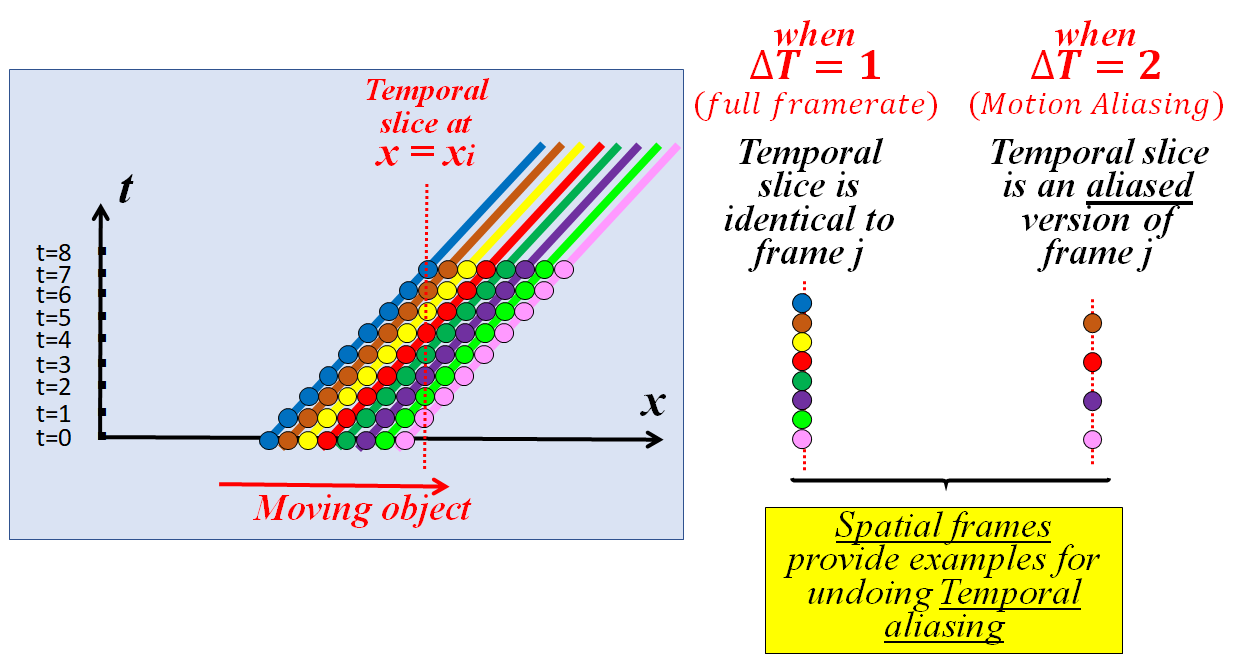}
    \caption{
\textbf{1D illustration of ``across dimension" recurrence.}
\emph{A 1D object moves to the right. When properly sampled in time ($\Delta$T=1), temporal slices are similar to spatial slices (1D ``frames''). However, when the temporal sampling rate is too low ($\Delta$T=2), temporal slices are undersampled  (aliased) versions of the spatial slices. Thus, spatial frames provide examples for undoing the temporal aliasing.  (See text for more details.)
\vspace*{-0.5cm}}
 }
\label{fig:2dBall_Illustration}
\end{figure}
%
Fig.~\ref{fig:2dBall_Illustration} explains this phenomenon in a simplified ``flat'' world.  A  1D object moves horizontally to the right with constant speed.
The 2D space-time plane here (xt), is equivalent to the 3D space-time video volume (xyt) in the general case. If we look at a specific point $x$$=$$x_i$, the entire object passes through this location over time. Hence looking at the temporal slice through $x$$=$$x_i$ (here the slices are 1D lines), we can see the entire object emerging in that temporal slice. The resolution of the 1D temporal slice depends on the object's speed compared to the framerate. For example, if the object's speed is $1~pixel/second$,
then taking frames every $1 second$ ($\Delta t = 1$) will show the entire object in the 1D temporal slice at $x$$=$$x_i$ . However, if we sample slower in time (which is equivalent to a faster motion with the same framerate), the temporal slice at $x$$=$$x_i$  will now display an \underline{aliased} version of the object (Fig.~\ref{fig:2dBall_Illustration}, on the right).  In other words, the spatial frame at $t$$=$$t_0$  is a high-resolution version of the aliased temporal slice at $x$$=$$x_i$. 
The full-resolution \emph{spatial} frame at $t$$=$$t_0$  thus teaches us how to \emph{undo} the motion (temporal) aliasing of the \emph{temporal} slice  at $x$$=$$x_i$ .

The same applies to the 3D video case. When a 2D object moves horizontally with constant speed, the y-t slice will contain a downscaled version of that object. The higher-resolution x-y frames teach how to undo that temporal aliasing in the y-t slice.  Obviously, objects in natural videos do not necessarily move in a constant speed. This however is not a problem, since our network resolves only small space-time video patches, relying on the speed  being constant only \emph{locally} in space and time (e.g., within a 5$\times$5$\times$3 space-time patch).  



Shahar et. al~\cite{Shahar2011STSR} showed that small 3D ST (Space-Time)  patches tend to recur in a video sequence, within/across multiple spatio-temporal scales of the video. We refer to these recurrences as \emph{`recurrence within the same dimension'} (i.e., no swap between the axes of the video volume).
Patch \emph{`recurrence across dimensions'} provides {additional} high-quality internal examples for temporal-SR. 
This is used \emph{in addition} to the patch recurrence within the same dimension. 
%

\begin{figure}[!t]
    \centering
    \includegraphics[width=0.8\linewidth]{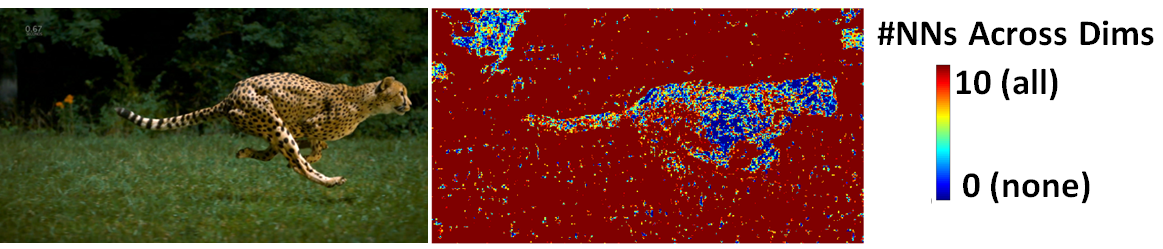}
    \caption{\textbf{Similarity Across-Dimensions:} 
   \emph{ \small  Nearest-Neighbor (NN) heat map indicating the percent of best patch matches (out of 10), found ``across-dimensions'' vs. ``within the same dimension''. 
    Red color indicates patches for which all 10 best matches were found ``{across dimension}'';  Blue  indicates patches for which all 10 best matches were found ``{within the same dimension}''. 
    As can be seen, a significant portion of the ST-patches found their best matches across dimensions. }
    \label{fig:across-dims_111}
    \vspace*{-.5cm}}
\end{figure}

Fig.~\ref{fig:across-dims_111} visually conveys the strength of {`recurrence across dimensions'} of small  ST-patches in the Cheetah video, compared to their recurrence within the same dimension. Each 5$\times$5$\times$3 patch in the original video searched for its top 10 approximate nearest-neighbors (using Patch-Match~\cite{barnes2010generalized}). 
These best matches were searched in various scales, both within the same dimension (in the original video orientation, $xyt$), and across dimensions ($tyx$, by flipping the video volume so that the x-axis becomes the new t-axis).  The colors indicate how many of these best matches were found across-dimension. Red color ($100\%$) indicates patches for which all 10 best matches were found \emph{across dimension};  Blue ($0\%$) indicates patches for which all 10 best matches were found \emph{within the same dimension}. The figure illustrates that a significant portion of the ST-patches found their best matches across dimensions (showing here one slice of the video volume). These tend to be patches with large motions -- the background in this video (note that the background moves very fast, due to the fast camera motion which tracks the cheetah). Indeed, as explained above, patches with large motions can benefit the most from using the cross-dimension examples.

Both of these types of patch recurrences (within and across dimensions) are used to perform \emph{Zero-Shot Temporal-SR}  from a single video. Sec.~\ref{sec:generatingExamples} explains how to exploit these internal ST-patch recurrences to generate training examples from the input video alone. This allows to increase the framerate while undoing both motion blur and motion aliasing, by training a light \emph{video-specific CNN}.

Fig.~\ref{fig:glass_slices} shows that patch recurrence across dimensions applies not only to simple linear motions, but also in videos with very complex motions. A ball falling into a liquid filled glass was recorded with a circuiting slow-motion (high framerate) camera. We can see x-t and y-t slices from that video contain similar patches as in the original x-y frames. Had this scene been recorded with a regular (low-framerate) camera, the video frames would have provided high-resolution examples for the lower temporal resolution of the x-t and y-t slices.

\section{Generating an Internal Training Set}
\label{sec:generatingExamples}
\emph{Low-temporal-resolution} (LTR) \& \emph{High-temporal-resolution} (HTR) pairs of examples are extracted directly from the input video, giving rise to self-supervised training.
These example pairs are used to train a relatively shallow fully convolutional network, which learns to increase the temporal resolution of the ST-patches \emph{of this specific video}. 
Once trained, this video-specific CNN is applied to the input video, to generate a HTR output. 

The rationale  is: Small ST-patches in the input video recur in different space-time scales and different dimensions of the input video. Therefore, the same network that is trained to increase the temporal resolution of these ST-patches in other scales/dimensions of the video, will also be good for increasing their temporal-resolution in the \emph{input} video itself 
(analogous to ZSSR in images~\cite{ZSSR_a}).

The creation of  relevant training examples is thus crucial to the success of the learning process. 
In order for the CNN to generalize well to the input video, 
the LTR-HTR training examples should 
\emph{bear resemblance} and have similar statistics of ST-patches as in the input video and its (unknown) HTR version. This process is explained next.
%
\\

\begin{figure}[!t]
    \centering
    \includegraphics[width=0.8\linewidth]{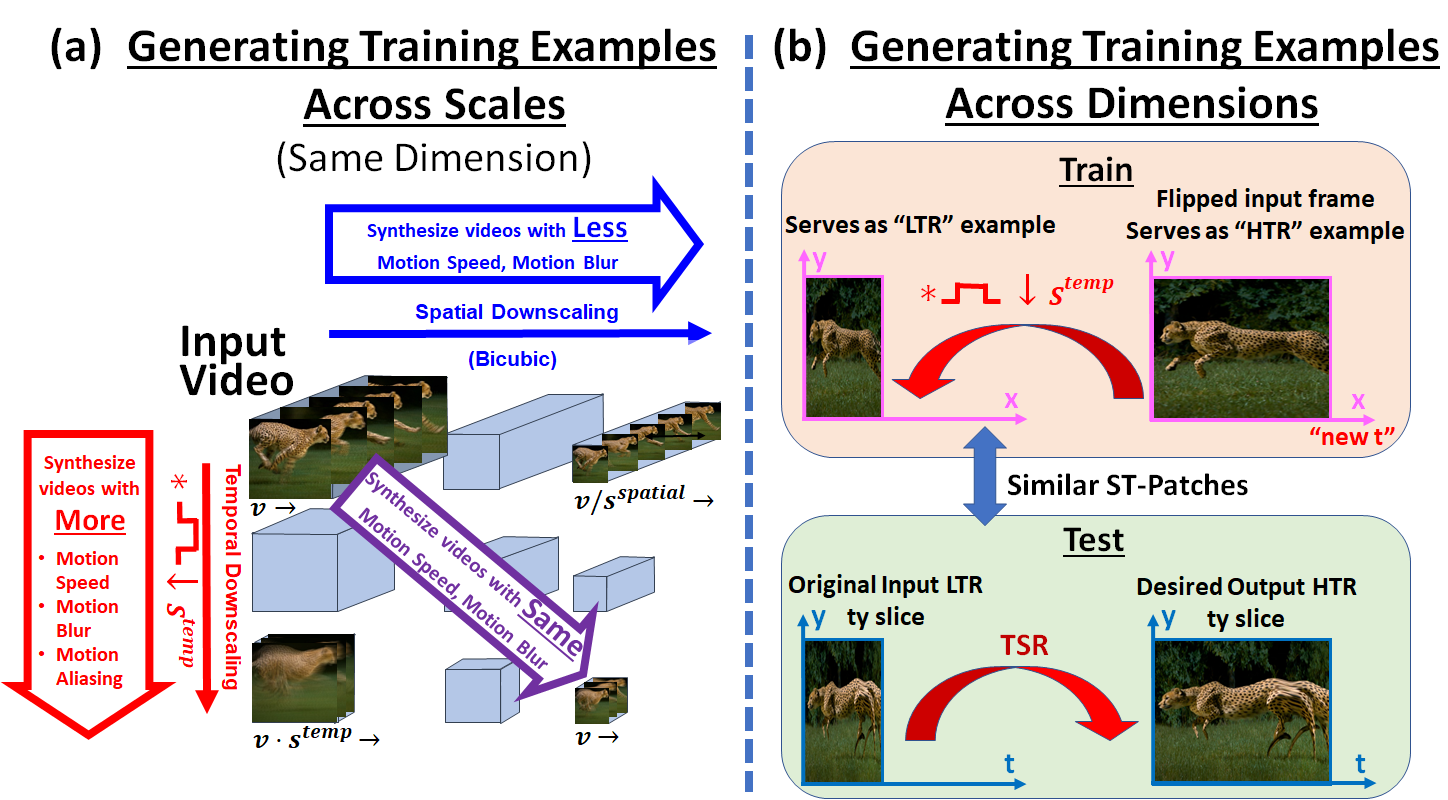}
    \caption{\textbf{Generating Internal Training Set.} 
 \emph{\small (a) Different combinations of spatio-temporal scales provide a variety
of speeds, sizes, different degrees of motion blur \& aliasing.
This generates a variety of LTR example videos,  for which their corresponding ground-truth HTR videos are known (the space-time volumes just above them).
(b) The xy video frames provide high-resolution examples for the ty and xt slices. Training examples can therefore be generated from these spatial frames, showing how to increase the temporal resolution.
\textbf{Please see the video in our \href{http://www.wisdom.weizmann.ac.il/\~vision/DeepTemporalSR}{project website} for a visual explanation and demonstration of those internal augmentations}.\vspace*{-0cm}}
    \label{fig:Training_set}}
    \vspace*{-0.45cm}
\end{figure}

\noindent\textbf{3.1~Example Pairs from ``Same Dimension'': }
%
\noindent The first type of training examples makes use of similarity of small ST-patches across spatio-temporal scales of the video. 
As was observed in~\cite{Shahar2011STSR}, and shown in Fig.~\ref{fig:Training_set}.a: \ \

\begin{itemize}[noitemsep,nolistsep,leftmargin=*,label=$\bullet$]
\item
Downscaling the video frames spatially (e.g., using bicubic downscaling), causes edges to appear sharper and move slower (in pixels/frame). This generates ST-patches with \emph{\underline{higher} temporal resolution}.
\item
Blurring and sub-sampling a video in time (i.e., reducing the framerate and increasing the ``exposure-time'' by averaging frames), causes an increase in speed, blur, and motion aliasing. This generates ST-patches with \emph{\underline{lower} temporal resolution}.
Since the ``exposure-time'' is  a highly \textbf{non-ideal LPF} (its temporal support is $\leq$ than the gap between 2 frames), such temporal coarsening introduces additional motion aliasing.

\item
Donwscaling by the same scale-factor both in space and in time (the diagonal arrow in Fig.~\ref{fig:Training_set}.a), preserves the same amount of speed and blur. This generates ST-patches with \emph{\underline{same} temporal resolution}.
\end{itemize}
%

%
\noindent 
Different combinations of spatio-temporal scales provide a variety
of speeds, sizes, different degrees of motion blur and different degrees of motion aliasing.
In particular, downscaling by the same scale-factor in space and in time (the diagonal arrow in Fig.~\ref{fig:Training_set}.a), generates a variety of LTR videos, whose ST-patches are similar to those in the LTR input video, but for which their corresponding ground-truth HTR videos are known (the corresponding space-time volumes just above them  in the space-time pyramid of Fig.~\ref{fig:Training_set}.a).

Moreover, if the same object moves at different speeds in different parts of the video (such as in the rotating fan/wheel videos in \href{http://www.wisdom.weizmann.ac.il/\~vision/DeepTemporalSR}{project website}),  the slow part of the motion provides examples how to undo the motion blur and aliasing  in faster parts of the video. Such LTR-HTR example pairs are obtained from the bottom-left part of the space-time pyramid (below the diagonal in Fig.~\ref{fig:Training_set}.a).


To further enrich the training-set with a variety of examples, we apply additional augmentations to the input video. These include mirror flips, rotations by $90^{\circ}, 180^{\circ},270^{\circ}$, as well as flipping the video in time. This is useful especially in the presence of chaotic non-rigid motions.
\\

\noindent\textbf{3.2~Example Pairs ``Across Dimensions'': }
%
%
\noindent In order to make use of the similarity between small ST-patches across dimensions (see Sec.~\ref{sec:across_dim}), we create additional training examples by rotating the 3D video volume -- i.e., swapping the spatial and temporal dimensions of the video. Such swaps are applied to a variety of spatially (bicubically) downscaled versions of the input video. Once swapped, a variety of 1D temporal-downscalings (temporal rect) are applied to the \emph{new} ``temporal'' dimension (originally the x-axis or y-axis).
The pair of volumes before and after such ``temporal'' downscaling form our training pairs.

While at test time the network is applied to the input video in its original orientation (i.e., TSR is performed along the original t-axis), training the network on ST-patches with similarity across dimensions creates a richer training set and improves our results.
Here  too, data augmentations are helpful (mirror flips, rotations, etc.). For example, if an object moves to the right (as in the Cheetah video), the  y-t slices will bare resemblance to \emph{mirror-reflected} versions of the original x-y frames (e.g., see the cheetah slices in Fig.~\ref{fig:All_slices}).

In our current implementation, we use both types of training examples (`within-dimension' and `across-dimensions'), typically with equal probability.
Our experiments have shown that in most videos, using both types of training examples is superior to using only one type 
(see also ablation study in Sec.~\ref{sec:experiments}).
\section{`Zero-Shot' Temporal-SR -- The Algorithm}

The repetition of  small ST-patches inside the input video (aross scales and across dimensions), provide ample data for training. Such an internal training-set concisely captures the characteristic statistics of the given input video: its local spatial appearances, scene dynamics, motion speeds, etc.
Moreover, such an internal training-set has relatively few ``distracting" examples which are irrelevant to the specific task at hand.
This is in stark contrast to the external training paradigm, where the vast majority of the training examples are irrelevant, and may even be harmful, for performing inference on a specific given video.
This high quality training allows us to perform true TSR using a simple conv net without any bells and whistles; our model has no motion~estimation nor optical~flow components, nor does it use any complicated building blocks. \\

\begin{figure}[!t]
    \centering
    \includegraphics[width=0.8\linewidth]{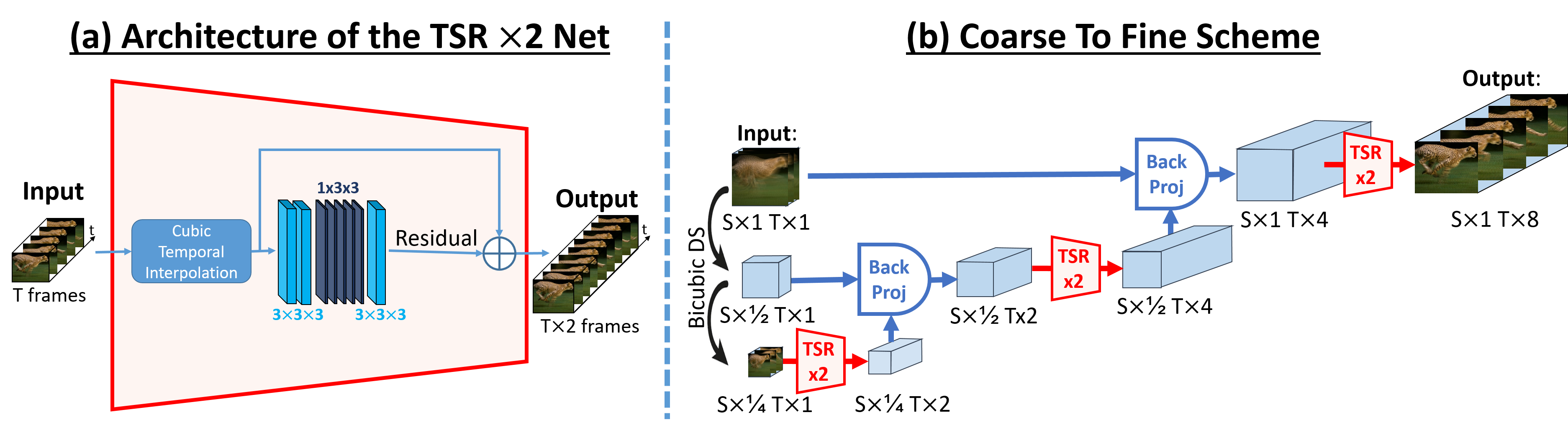}
    \caption{
       \textbf{Coarse to fine scheme and Architecture.} \textbf{\emph{(see text for details).}}
        \vspace*{-0.5cm}}
    \label{fig:Architecture}
\end{figure}

\noindent
\textbf{4.1 Architecture:} \
\noindent A fully Convolutional Neural Network (CNN)  efficiently calculates its output patch by patch. 
Each output pixel is a result of a calculation over a patch in the input video.
The size of that patch is determined by the effective receptive field of the net \cite{effective-receptive-field-2016}. 
Patch recurrence across scales and dimensions holds best for relatively small patches, hence we need to ascertain that the receptive field of our model is relatively small in size. 
Keeping our network and filters small (eight 3D conv layers, some  with 3$\times$3$\times$3 filters and some  with 1$\times$3$\times$3, all with stride 1), we ensure working on small patches as required. 
%
Each of our 8 conv layers has 128 channels, followed by a ReLU activation. 
The input to the network is a temporally interpolated video (simple cubic interpolation), and the network learns only the residual between the interpolated LTR video to the target HTR video. 
Fig.~\ref{fig:Architecture}.a provides a detailed description of our model.

At each iteration, a  36$\times$36$\times$16 space-time video crop   is randomly selected from the various internal augmentations (Sec.~\ref{sec:generatingExamples}). A crop is selected with probability proportional to its mean intensity gradient magnitude. This crop forms a HTR (High Temporal Resolution)  example. It is then blurred and subsampled by a factor of 2 \emph{in time}, to generate an internal  LTR-HTR training pair. 
%

An $\ell_2$ loss is computed on the recovered space-time outputs. 
We use an ADAM optimizer \cite{adam}.
The learning rate is initially set to $10^{-4}$, and is adaptively decreased according to the training
procedure proposed in~\cite{ZSSR_a}.
The training stops when the learning rate reaches  $10^{-6}$.

The advantage of  \emph{video-specific} internal training is the adaptation of the network to the specific  data at hand. The downside of  such Internal-Learning is that it requires training the network from scratch for each new input video. 
Our network requires $\sim$2 hours training time per video on a single Nvidia V100 GPU. 
%
Once trained, inference time at 720$\times$1280 spatial resolution is $\sim1.7 sec/frame$.\\ 


\noindent\textbf{4.2 \ Coarse-to-Fine Scheme \ (in Space \& in Time):} \ 
\noindent Temporal-SR becomes complex when there are large motions and severe blur. 
As shown in Fig.~\ref{fig:Training_set}.a, spatially downscaling the video results in smaller motions and less motion blur. Denoting the input video resolution by \SxTx{1}{1}, our goal is to recover a video with $\times 8$ higher temporal resolution: \SxTx{1}{8}. 
To perform our temporal-SR we use a coarse-to-fine approach (Fig.~\ref{fig:Architecture}.b). 

We start by training our network on a spatially downscaled version of the input video (typically \SxTx{1/8}{1}, or \SxTx{1/4}{1} for spatially small videos). 
Fig.~\ref{fig:Architecture}.b details a coarse-to-fine upscaling scheme from \SxTx{1/4}{1}.The scheme to upscale from \SxTx{1/8}{1} includes an additional ``Back-Projection'' stage at the end.
The network trains on this small video, learning to increase its temporal resolution by a factor of 2. Once trained, the network is applied to \SxTx{1/8}{1} to generate \SxTx{1/8}{2}. 
We then use ``Back-Projection''
\footnote{Don't confuse ``Back-Projection''~\cite{backprojection} with ``backpropagation''~\cite{goodfellow2016deep}.}~\cite{backprojection} 
(both spatially  and temporally), 
to increase the spatial resolution of the video by a factor of 2, resulting in \SxTx{1/4}{2}. 
The spatial Back-Projection guarantees the \emph{spatial} (bicubic) consistency of the resulting \SxTx{1/4}{2} with the spatially smaller \SxTx{1/8}{2}, and its \emph{temporal} (rect) consistency with the temporally coarser \SxTx{1/4}{1}.

Now, since we increased both the spatial and temporal resolutions by the same factor ($\times 2$), the motion sizes and blurs in \SxTx{1/4}{2} remain similar in their characteristics to those in \SxTx{1/8}{1}. This allows us to apply the same network again, as-is, to reach a higher temporal resolution: \SxTx{1/4}{4}. 
We iterate through these two steps: increasing temporal resolution using our network, and subsequently increasing the spatial resolution via spatio-temporal Back-Projection, going up the diagonal in Fig.~\ref{fig:Training_set}.a, until we  reach the goal resolution of \SxTx{1}{8}.

The recurring use of TSRx2 and "Back-Projection" accumulates errors.
Fine-tuning at each scale is likely to improve our results, and also provide a richer set of training examples as we go up the coarse-to-fine scales. 
However, fine-tuning was not used in our current reported results due to the tradeoff in runtime.

\section{Experiments \& Results}
\label{sec:experiments}
%
True TSR (as opposed to simple frame interpolation) is mostly in-need when temporal information in the video is severely under-sampled and lost, resulting in motion aliasing. Similarly, very fast  stochastic motions recorded within a long exposure time result in unrecognizable objects.
To the best of our knowledge, a  dataset of such low-quality (LTR) videos of complex dynamic scenes, along with their ``ground truth'' HTR videos, is not publicly available. Note that these are very different from datasets used by frame-interpolation methods
 (e.g.,~\cite{xue2019video,galoogahi2017needforspeed,soomro2012dataset,baker2011database,bao2019memc}), since these do not exhibit motion blur or motion aliasing, and hence are irrelevant for the task of TSR.

%
We therefore curated a challenging dataset of 25 LTR videos of very complex fast dynamic scenes, ``recorded'' with a `slow' (30 fps) video camera  \emph{with full inter-frame exposure time}.
The dataset was generated from real complex videos recorded with 
high speed cameras (mostly 240 fps). The LTR videos were generated from our HTR `ground-truth' videos by \emph{blurring and sub-sampling them in time} 
(averaging every 8 frames). 
%
%
Since these 25 videos are quite long, they provide ample data (a very large number of frames) to 
compare and 
evaluate on.
We further split our LTR dataset into 2 groups: 
(i)~13 extremely challenging videos, not only with severe motion blur, but also with severe motion aliasing and/or  complex highly non-rigid motions (e.g., splashing water, flickerig fire); \ 
(ii)~12 less challenging videos, still with sever motion blur, but mostly rigid motions.



Fig.~\ref{fig:visual-results} displays a few such examples for TSR$\times$8. 
We compared our results (both visually and numerically) to the leading methods in the field (DAIN~\cite{DAIN},  NVIDIA SloMo~\cite{nvidia_slomo}, Flawless\cite{flawless}). 
As can be seen, complex dynamic scenes pose a challenge to all methods. Moreover, the rotating fan/wheel, which induce severe motion blur and severe motion aliasing, cannot be resolved by any of these methods. Not only are the recovered  frames extremely distorted and blurry (as seen in Fig.~\ref{fig:visual-results}), they all recover a false direction of motion (counter-clockwise rotation), and with a wrong rotation speed.
\emph{\textbf{The reader is urged to view the videos in our \href{http://www.wisdom.weizmann.ac.il/\~vision/DeepTemporalSR}{project website} in order to see these strong aliasing effects.}}
Table~\ref{tab:benchmark} provides quantitative comparisons of all methods on our dataset --  compared using PSNR, structural similarity (SSIM), and a perceptual measure (LPIPS\cite{lpips2018}).
The full table of all 25 videos is found in the \href{http://www.wisdom.weizmann.ac.il/\~vision/DeepTemporalSR}{project website}.
Since Flawless is restricted to $\times$10 temporal expansion (as opposed to the $\times$8 of all other methods), we ran it in a slightly different setting, so that their results could be compared to the same ground truth. 
{Although most closely related to our work, we could not  compare to~\cite{Shahar2011STSR}, due to its outdated  software. Moreover, our end-to-end method is currently adapted to TSRx8, whereas their few published results are TSRx2 and TSRx4, hence we could not visually compare to them either (our TSRx2 network can currently train only on small (coarse) spatial video scales, whereas~\cite{Shahar2011STSR} applies SRx2 to their fine spatial scale.}

The results in Table~\ref{tab:benchmark} indicate that sophisticated frame-interpolation methods (DAIN \cite{DAIN}, NVIDIA SloMo~\cite{nvidia_slomo}) are not adequate for the task of TSR, and are {significantly} inferior (-1 dB)  on LTR videos compared to dedicated 
TSR methods (Ours and Flawless~\cite{flawless}). 
{In fact, they are not much better (+0.5 dB) than plain intensity-based linear interpolation on those videos}.
Flawless and Ours provide comparable quantitative results on the dataset, even though Flawless is a {pre-trained supervised} method, whereas Ours is \emph{unsupervised} and requires no prior training examples. 
Moreover, {on the subset of extremely challenging videos (highly complex non-rigid motions), our Zero-Shot TSR  outperforms the externally trained Flawless~\cite{flawless}}. 
We attribute this to the fact that it is practically infeasible to generate an exhaustive enough  external training set to  cover the variety of all possible  \emph{non-rigid} motions. 
In contrast, highly relevant \emph{video-specific} training examples are found internally, inside the LTR input video itself.  
%
%


Since rigid motions are easier to model and capture in an external training set, Flawless provided high-quality results (better than ours) on the videos which are dominated by rigid motions. However, even in those videos, when focusing on the areas with non-rigid motions, our method visually outperforms the externally trained Flawless. While these non-rigid areas are smaller in those videos (hence have negligible effect on PSNR), they often tend to be the salient and  more interesting regions in the frame. Such examples are found in Fig.~\ref{fig:visual-results} (e.g., the billiard-ball and hoola-hoop examples), and in the videos in  \href{http://www.wisdom.weizmann.ac.il/\~vision/DeepTemporalSR}{project website}. \\

\begin{table}[!t]
    \centering
    \begin{tabular}{|r||r||c|c|c|c|c|}
        \hline
        & &\quad \textcolor{red}{\textbf{Ours}} \quad\ & Flawless \cite{flawless} & DAIN \cite{DAIN} & Nvidia  & linear  \\
        & &                                            &                          &                  & SloMo \cite{nvidia_slomo} & interp. \\
         \hline 
         \hline
         & PSNR [dB] \hspace*{0.01cm}& \textbf{28.63} & 28.22  &  27.29 & 27.23 & 26.79 \\
        \textbf{Entire} \hspace*{0.32cm}  & SSIM \hspace*{0.68cm} & 0.917 &  \textbf{0.918}  &  0.903 & 0.901 & 0.895 \\
        \textbf{Dataset} \hspace*{0.25cm} & LPIPS$^\dagger$\cite{lpips2018} & \textbf{0.174} &  \textbf{0.174}  &  0.214 & 0.214 & 0.231 \\
         \hline
         \hline
         & PSNR [dB] \hspace*{0.01cm} & \textbf{28.58} &  27.58  &  27.09 & 27.03 & 26.99 \\
        \textbf{Challenging}  & SSIM\hspace*{0.75cm} & \textbf{0.929} &  0.918  &  0.909 & 0.906 & 0.907 \\
         \textbf{Videos} \hspace*{0.28cm} & LPIPS$^\dagger$\cite{lpips2018} & \textbf{0.161} &  0.188  &  0.208 & 0.205 & 0.212 \\         
         \hline   
    \multicolumn{7}{l}
    {\footnotesize \textcolor{blue}{$^\dagger$ LPIPS (percetual \emph{distance}) -- lower is better. \  PSNR and SSIM -- higher is better.}}
     \vspace*{0.1cm}
    \end{tabular}

    \caption{\textbf{Comparing TSR$\times$8  results on our dataset.} 
     \emph{\small When applied to challenging videos with severe motion blur and motion aliasing, sophisticated frame upsampling methods (Nvidia SlowMo and DAIN) score significantly lower. However, even methods trained to overcome such challenges (e.g., Flawless), but were pre-trained on an external dataset, struggle to compete on videos that deviate from the typical motions and dynamic behaviors they were trained on. 
    \vspace*{-0.35cm}
}}
    \label{tab:benchmark}
    \vspace*{-0.75cm}
 \end{table}

\noindent
\textbf {Ablation Study:}
One of the important findings of this paper is the strong patch recurrence \emph{across-dimensions}, and its implication on extracting useful internal training examples for TSR. To examine the power of such cross-dimension augmentations, we conducted an ablation study. Table~\ref{tab:ablation} compares the performance of our network when: (i)~Training only on examples from \emph{same-dimension} (`Within'); (ii) Training only on examples \emph{across-dimensions} (`Across'); (iii) Training each video on its best configuration -- `within',  `across', or on both. 

Since our atomic TSRx2 network is trained only on a coarse spatial scale of the video, we performed the ablation study at that scale (hence the differences between the values in Tables~\ref{tab:benchmark} and~\ref{tab:ablation}). This allowed us to isolate purely the effects of the choice of augmentations on the training, without the distracting effects of the subsequent spatial and temporal Back-Projection steps.
Table~\ref{tab:ablation} indicates that, on the average, the cross-dimension augmentations are more informative than the within (same-dimension) augmentations. 
However, since different videos have different preferences, training each video with its best within and/or across configuration provides 
an additional overall improvement in PSNR, SSIM and LPIPS (improvements are shown in blue parentheses  in Table~\ref{tab:ablation}).

This suggests that each video should ideally be paired with its best training configuration -- a viable option with Internal training. 
For example, our video-specific ablation study indicated that videos with large uniform  motions tend to benefit significantly more from cross-dimension training examples (e.g., the falling diamonds video in Fig.~\ref{fig:visual-results} and in the \href{http://www.wisdom.weizmann.ac.il/\~vision/DeepTemporalSR}{project website}). In contrast,  videos with gradually varying speeds or with rotating motions tend to benefit from within-dimension examples (e.g., the rotating fan video in Fig.~\ref{fig:visual-results} and in the \href{http://www.wisdom.weizmann.ac.il/\~vision/DeepTemporalSR}{project website}). Such general video-specific preferences can be estimated per video by using very crude (even inaccurate) optical-flow estimation at very coarse spatial scales of the video. This is part of our  future work.
In the meantime, our default configuration randomly samples  augmentations from both `within' (same-dimension) and `across-dimensions'. \\
%

\begin{table}[!t]
    \centering
    \begin{tabular}{|r||c|c|c|}
        \hline
        & Only Within & Only Across & Best of all configurations \\
         \hline 
         \hline
          PSNR [dB] \hspace*{0.02cm} &  33.96  &  34.25 \improve{0.28}  & 34.33 \improve{0.37}\\
         SSIM \hspace*{0.75cm} &  0.962  &  0.964 \improve{0.002}  & 0.965 \improve{0.003}\\
          LPIPS$^\dagger$~\cite{lpips2018} &  0.035  &  0.033 \improve{0.002}  & 0.032 \improve{0.003}\\
         \hline
    \multicolumn{4}{l}
    {\footnotesize \textcolor{blue}{$^\dagger$ LPIPS (percetual \emph{distance}) -- lower is better. \  PSNR and SSIM -- higher is better.\vspace*{0.1cm}}}
    \end{tabular}
    
    \caption{\textbf{Ablation study: `Within' vs. `Across' examples.}
     \emph{\small Average results of our atomic TSRx2 network,  when trained on examples extracted from: (i)~\underline{same-dimension} only (`Within'); (ii)~\underline{across-dimensions} only  (`Across'); (iii)~\underline{best configuration} for each video  (`within',  `across',  or both). 
The ablation results indicate that on average, the cross-dimension augmentations are more informative than the within (same-dimension) augmentations, leading to an overall improvement in PSNR, SSIM and LPIPS  (improvements shown in blue parentheses). 
However, since different videos have different preferences, training each video with its best `within' and/or `across' configuration can provide an additional overall improvement in all 3 measures.
\vspace*{-0.35cm}
}}
    \label{tab:ablation}
\vspace*{-0.25cm}
\end{table}

\noindent
\textbf{6. Conclusion} \\
We present an approach for Zero-Shot Temporal-SR, which requires no training examples other than the input test video. 
Training examples are extracted from coarser spatio-temporal scales of the input video, as well as from other video dimensions (by swapping space and time).
Internal-Training adapts to the data-specific statistics of the input  data. It is therefore more adapted to cope  with new challenging (never-before-seen) data.
Our approach can resolve motion blur and motion aliasing in very complex dynamic scenes,   
surpassing previous supervised methods trained on external video datasets.\\

\clearpage
\noindent
\textbf{Acknowledgments:} \ 
Thanks to Ben Feinstein for his \mbox{invaluable} help in getting the GPUs to run  smoothly and  efficiently.
This project received funding from the European Research Council (ERC) Horizon 2020, grant No 788535,
from the Carolito Stiftung and by grant from D. Dan and Betty Kahn Foundation.
Dr Bagon is a Robin Chemers Neustein AI Fellow.


{\small
\bibliographystyle{splncs04}
\bibliography{st-zssr.bib}
}

\end{document}